\documentclass[letterpaper, 10 pt, conference]{ieeeconf}

\IEEEoverridecommandlockouts
\usepackage{times}
\usepackage{cuted}
\usepackage{caption}
\usepackage{multicol}
\usepackage[bookmarks=true]{hyperref}
\usepackage[dvipsnames]{xcolor}

\ifdefined\showrevisions
  \newcommand{\rev}[1]{\textcolor{blue}{#1}}
\else
  \newcommand{\rev}[1]{#1}
\fi
\usepackage{amsmath,amssymb}
\usepackage{bm}
\usepackage{graphicx}
\usepackage{siunitx}
\usepackage{tikz}
\usetikzlibrary{arrows.meta,fit,positioning}
\usepackage{pgfplots}
\usepgfplotslibrary{groupplots}
\pgfplotsset{compat=1.18}
\usepackage{adjustbox}
\usepackage{listings}
\definecolor{codebg}{RGB}{248,248,248}
\definecolor{codegreen}{RGB}{52,140,49}
\definecolor{codeblue}{RGB}{0,70,170}
\definecolor{codepurple}{RGB}{150,0,150}
\definecolor{codestring}{RGB}{150,70,20}
\definecolor{codegray}{RGB}{105,105,105}
\definecolor{mjorbitred}{rgb}{0.55,0.1,0.1}
\definecolor{plotDarkBlue}{HTML}{003A7D}
\definecolor{plotMedBlue}{HTML}{008DFF}
\definecolor{plotPink}{HTML}{FF73B6}
\definecolor{plotPurple}{HTML}{C701FF}
\definecolor{plotGreen}{HTML}{4ECB8D}
\definecolor{plotOrange}{HTML}{FF9D3A}
\definecolor{plotYellow}{HTML}{F9E858}
\definecolor{plotRed}{HTML}{D83034}
\newcommand{\R}{\mathbb{R}}
\newcommand{\mj}{MuJoCo}
\DeclareRobustCommand{\brandmono}[1]{{\normalfont\ttfamily\textcolor{mjorbitred}{#1}}}
\DeclareRobustCommand{\mjorbit}{\brandmono{mjorbit}}
\DeclareRobustCommand{\mjorbitwarp}{\brandmono{mjorbit-warp}}
\DeclareRobustCommand{\mjorbittitle}{\texttt{\textbf{\textcolor{black}{mjorbit}}}}

\IfFileExists{mjorbit_version_frozen.tex}{%

}{%
  \IfFileExists{mjorbit_version.tex}{%
    \input{mjorbit_version.tex}%
  }{%
  }%
}

\title{\LARGE \bf
{\Huge \mjorbittitle}:
A Simulation Framework for Space Robotics
}

\author{John Z. Zhang$^{1}$, Joris Verhagen$^{2}$, Fausto Vega$^{1}$, Patrick McKeen$^{1}$, and Zachary Manchester$^{1}$%
\thanks{$^{1}$ Department of Aeronautics and Astronautics, Massachusetts Institute of Technology. Correspondance to: {\tt\small jzhang3@mit.edu}}%
\thanks{$^{2}$ Division of Robotics, Perception and Learning, KTH Royal Institute of Technology. 
}%
}

\begin{document}

\maketitle
\thispagestyle{empty}
\pagestyle{empty}

\begin{abstract}
This paper presents a general framework for simulating multi-body space robots with contact. We bring efficient, large-scale robot simulation to in-space servicing, assembly, and manufacturing applications. First, we perform an empirical trade study of methods for coupling orbit propagation with existing robotics simulation frameworks. Next, we present \mjorbit{}, a general, flexible, and performant framework built on the MuJoCo engine widely used in robotics, to which we add key spacecraft dynamics, actuators, and sensors. We provide a low-latency C++ CPU backend and a high-throughput GPU backend behind a simple Python API. We demonstrate \mjorbit{} by solving several realistic on-orbit case studies with both model-predictive control and reinforcement learning. Open-source code and examples are available at:
\begin{center}
\url{https://johnzhang3.github.io/mjorbit/}
\end{center}
\end{abstract}

\section{Introduction}
\label{sec:introduction}

Contact-rich interaction is central to upcoming space missions, including on-orbit servicing, active debris removal, and in-space assembly. A robot may need to dock with a client spacecraft, grasp a tumbling non-cooperative target, manipulate an articulated payload, or stabilize an object after capture. These behaviors couple orbital motion, attitude dynamics, articulated multibody dynamics, actuation, sensing, and intermittent contact over long horizons. However, existing simulators tend to emphasize different parts of this problem: astrodynamics frameworks~\cite{kenneally2020basilisk} provide modular orbit, attitude, environmental, and flight-software simulation capabilities, but are typically missing multibody and contact dynamics modeling, and often don't support hardware acceleration. Robotics simulators~\cite{todorov2012mujoco, drake2019, howell2022dojo} provide efficient, increasingly GPU-accelerated implementations of articulated rigid-body dynamics and contact but do not model orbital mechanics and environmental effects on the spacecraft. Recent works have explored bridging these gaps~\cite{garciaBonilla2026articulated, schwartz2026smallsatsim, costi2025softnets}, yet a general, efficient framework for space robotics simulation is still missing.

\begin{figure}[t]
    \centering
    \begin{tikzpicture}[
        panel/.style={inner sep=0, outer sep=0},
        lbl/.style={
            anchor=north west, text=white, font=\bfseries\footnotesize,
            inner sep=1.6pt, fill=black, fill opacity=0.55, text opacity=1,
            rounded corners=1pt
        },
        node distance=0.8mm
    ]
        \node[panel] (a)
            {\includegraphics[width=0.495\columnwidth]{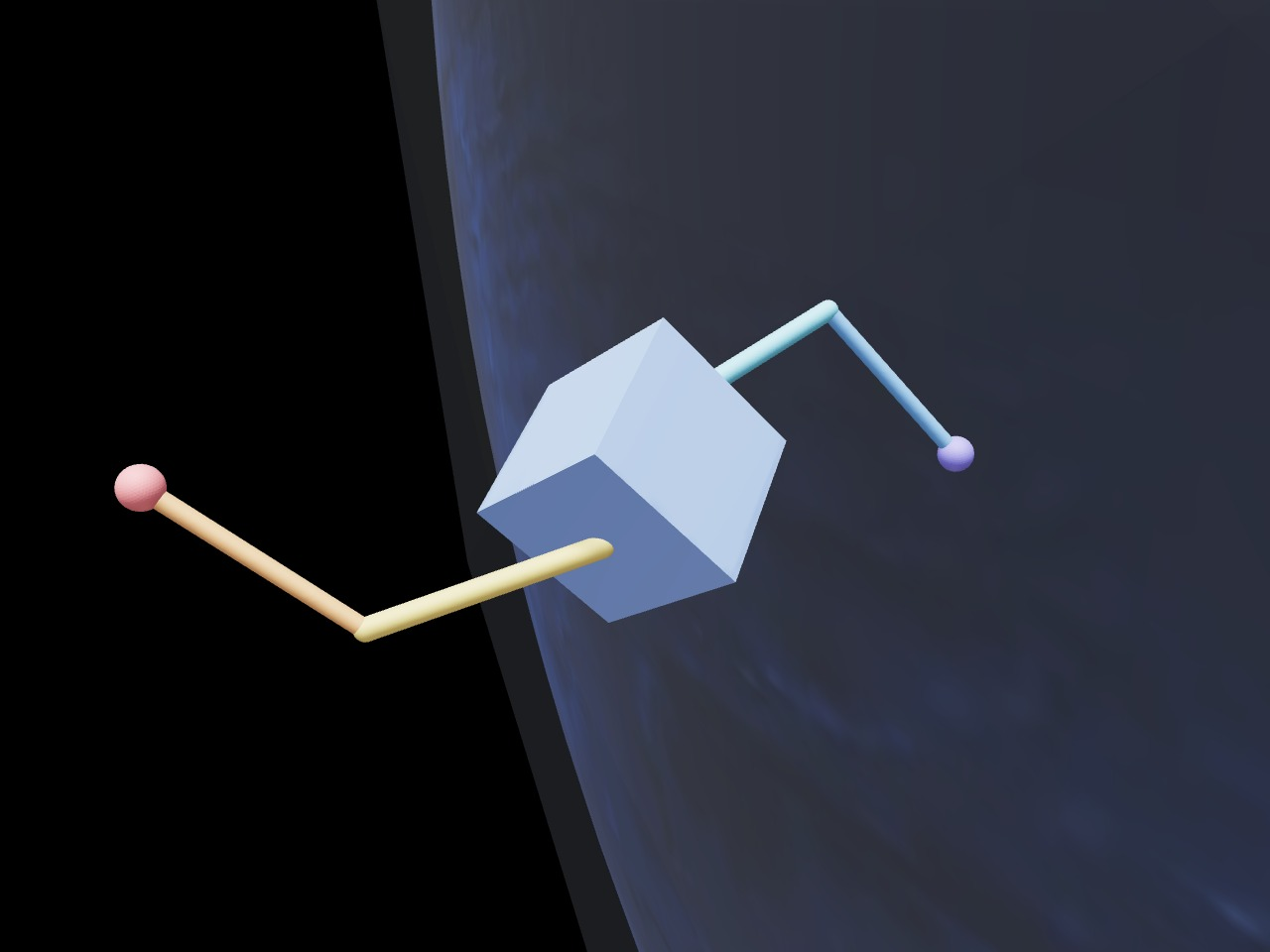}};
        \node[panel, right=of a] (b)
            {\includegraphics[width=0.495\columnwidth]{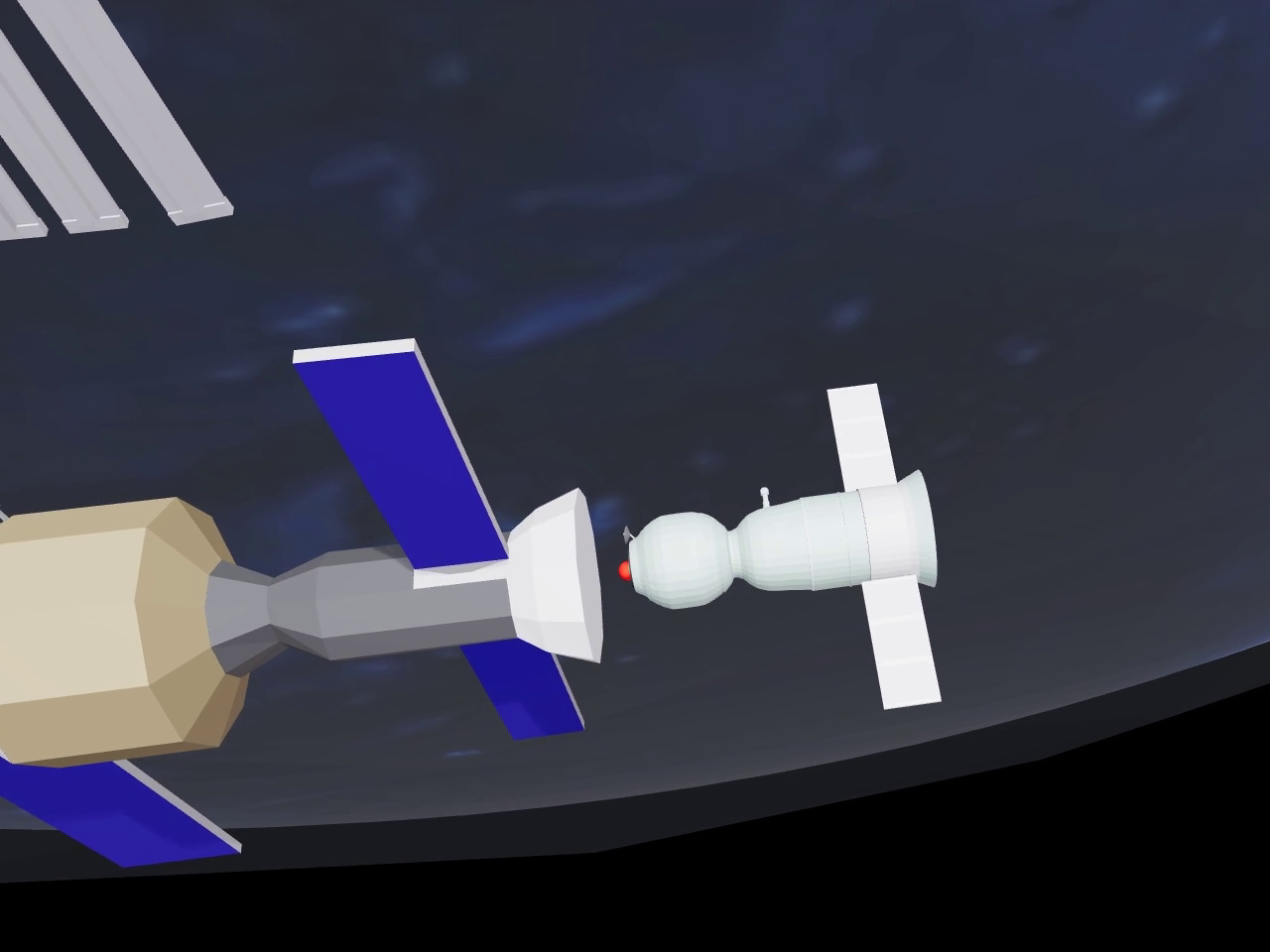}};
        \node[panel, below=of a] (c)
            {\includegraphics[width=0.495\columnwidth]{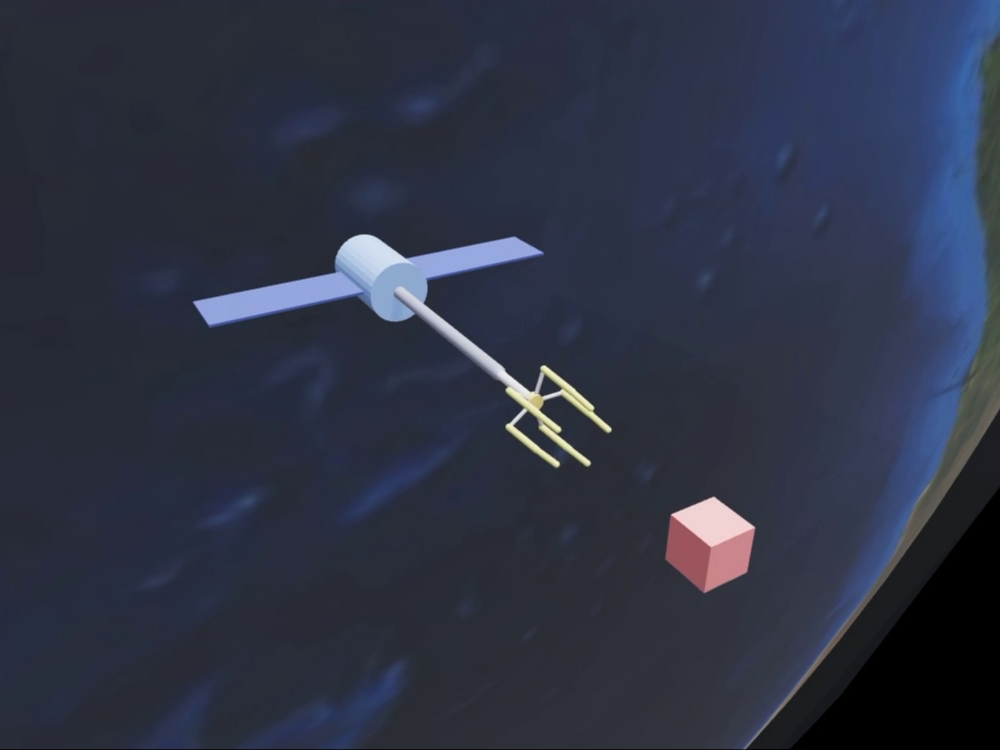}};
        \node[panel, right=of c] (d)
            {\includegraphics[width=0.495\columnwidth]{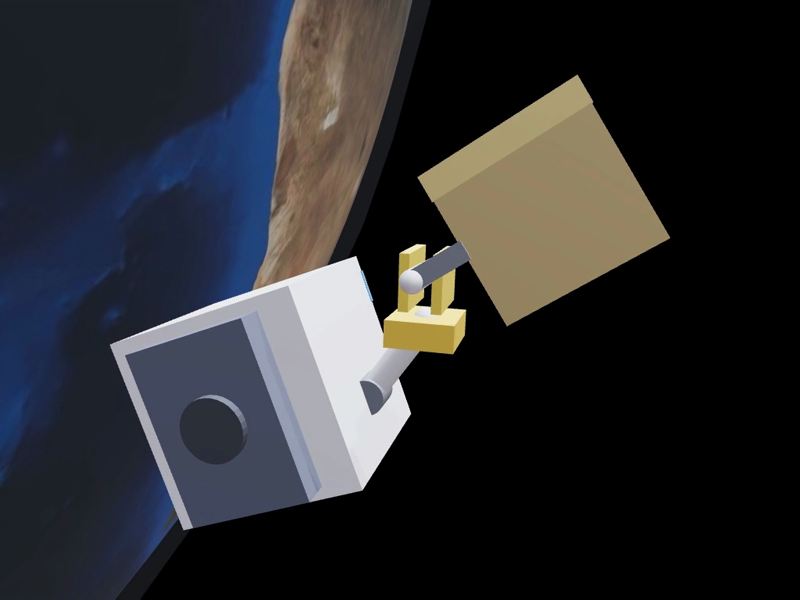}};
        \foreach \p/\t in {a/a, b/b, c/c, d/d}
            \node[lbl] at (\p.north west) {(\t)};
    \end{tikzpicture}
    \caption{Example control policies from \mjorbit{}: (a) attitude control using articulated arms, (b) autonomous rendezvous and docking, (c) grasping an orbit payload, and (d) Astrobee robot detumbles and grasps a free-floating cargo module.}
    \label{fig:examples}
    \vspace{-10pt}
\end{figure}

We bridge this gap with a general space-robotics simulation framework that provides accurate long-horizon physics and is tailored for policy learning and control. We analyze common choices when designing such a framework and demonstrate that our implementation is accurate both over long horizons and in single-precision arithmetic on GPUs~\cite{zakka2026mjlab}. Our framework, \mjorbit{}, builds on the MuJoCo physics engine~\cite{todorov2012mujoco} and its GPU-optimized variant, MJWarp~\cite{mujocoWarp2026}. We model orbital dynamics and perturbations --- including J2, atmospheric drag, solar radiation pressure, and magnetic torques. We also support common spacecraft actuators and sensors: thrusters, reaction wheels, control-moment gyros, magnetic torque coils, magnetometers, and sun sensors. To integrate with existing sim-to-real policy-learning~\cite{mittal2025isaacLab,zakka2026mjlab,rudin2022learningWalk} and control~\cite{howell2022predictiveSampling,alvarezPadilla2025wholeBodyMppi,zhang2025wholeBodyMpcMujoco,zhang2026sumo} pipelines, we adopt a MuJoCo-like API~\cite{todorov2012mujoco}.

Our specific contributions include:
\begin{itemize}
    \item A framework for coupling existing robotics simulation with orbital dynamics.
    \item A CPU implementation for low-latency physics simulation in C++ and a GPU implementation for high-throughput data generation.
    \item A set of examples using \mjorbit{} for real-time control and policy training.
\end{itemize}

The paper is organized as follows: Sec.~\ref{sec:related_work} and Sec.~\ref{sec:background} review related work and background, Sec.~\ref{sec:methods} studies frame and integrator choices under limited numerical precision, Sec.~\ref{sec:mjorbit} presents the \mjorbit{} framework, Sec.~\ref{sec:experiments} demonstrates it on control and policy-learning examples, and Sec.~\ref{sec:conclusion} concludes with limitations and future work.

\section{Related Work}\label{sec:related_work}
\subsection{Astrodynamics and Spacecraft Simulation}
Spacecraft simulation tools have traditionally focused on orbit propagation, attitude dynamics, environmental forces and torques, navigation sensors, and flight software. Tools like GMAT~\cite{nasaGmat}, Orekit~\cite{orekit}, and Basilisk~\cite{kenneally2020basilisk} provide mature support for trajectory design, orbit propagation, estimation, and operational analysis. Rigid-body ADCS simulators such as Generalized ADCS~\cite{scheuer2026generalizedAdcs} provide lightweight closed-loop orbit and attitude simulation, but do not target articulated contact dynamics. More generally, spacecraft simulators often assume a single rigid body or require specialized modeling for complex articulated mechanisms.

Several institutional simulators address richer multibody spacecraft dynamics. The Dshell-DARTS~\cite{garciaBonilla2025dshellDarts} toolkit developed at JPL supports aerospace and robotics simulation across orbital, interplanetary, atmospheric, and surface domains, including rigid and flexible multibody dynamics, gravity and atmosphere models, and collision modeling. MBJEOD integrates MBDyn multibody dynamics with JEOD orbital dynamics to support human-spaceflight scenarios such as free-flyer capture, maneuver, and release~\cite{sullivan2024mbjeod}. These systems show that combining orbital and multibody dynamics is feasible, but they are closed-source, oriented toward mission-specific engineering workflows, and not used as high-throughput engines for modern robotics.

An emerging trend in space robotics adopts existing robotics simulators and configures them for microgravity~\cite{schwartz2026smallsatsim}.
This preserves the articulated, flexible-body, and contact dynamics central to microgravity manipulation, tumbling-target capture and detumbling~\cite{wu2018captureContact,li2021combinedSpacecraft}, soft-net debris capture~\cite{costi2025softnets}, and adhesive grasping~\cite{jiang2017roboticdeviceusinggecko}.
However, while these simulators reasonably approximate microgravity
environments over short timescales (e.g. minutes), they do not model important environmental forces and torques that accumulate over orbital timescales (e.g. hours). 
This paper bridges this gap with an open-source, high-throughput, and long-term accurate simulation framework. Table~\ref{tab:simulator-comparison} summarizes the capabilities of existing space robot simulators.

\begin{table}[h]
    \centering
    \caption{Comparison of simulation for space robotics.}
    \label{tab:simulator-comparison}
    \scriptsize
    \renewcommand{\arraystretch}{1.08}
    \setlength{\tabcolsep}{2pt}
    \begin{tabular}{@{}p{0.29\columnwidth}*{5}{>{\centering\arraybackslash}p{0.118\columnwidth}}@{}}
        \hline
        Simulator & Orbit & Env. & CPU & GPU & M-body \\
        & Dynamics & Perturb. & Parallel & Parallel & Contact \\
        \hline
        Gen. ADCS \cite{scheuer2026generalizedAdcs}
            & $\checkmark$ & $\checkmark$ & $\times$ & $\times$ & $\times$ \\
        Basilisk \cite{kenneally2020basilisk,garciaBonilla2026articulated}
            & $\checkmark$ & $\checkmark$ & $\checkmark$ & $\times$ & $\checkmark$ \\
        Dshell-DARTS \cite{garciaBonilla2025dshellDarts}
            & $\checkmark$ & $\checkmark$ & $\times$ & $\times$ & $\checkmark$ \\
        SmallSatSim \cite{schwartz2026smallsatsim}
            & $\times$ & $\times$ & $\checkmark$ & $\checkmark$ & $\checkmark$ \\
        \textbf{\mjorbit{} (ours)}
            & $\checkmark$ & $\checkmark$ & $\checkmark$ & $\checkmark$ & $\checkmark$ \\
        \hline
    \end{tabular}
\end{table}
\begin{figure}[t]
    \centering
    \resizebox{0.99\columnwidth}{!}{\input{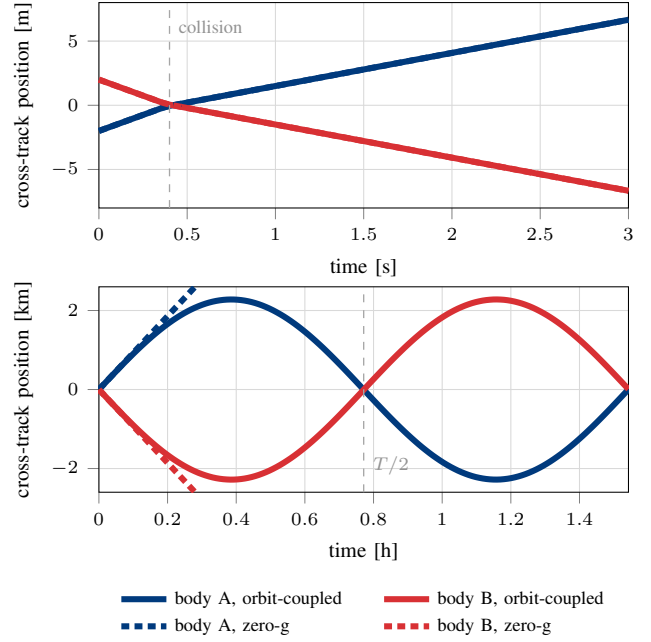}}
    \caption{Cross-track position of two free-floating boxes after impact, orbit-coupled (solid) vs.\ zero-g (dashed) from identical initial states. Top: the two agree across the contact event. Bottom: over one reference orbit, the orbit-coupled bodies oscillate boundedly and recross at $T/2$, while the zero-g simulation drifts without bound.}
    \label{fig:zero-g-comparison}
\end{figure}
\subsection{Robotics Simulation for Learning and Control}
Robotics simulators today build on decades of algorithmic literature for articulated rigid-body dynamics~\cite{luh1980online,featherstone2008rigidBody}, collision detection~\cite{gilbert1988fast,montaut2024gjkpp}, and contact mechanics~\cite{stewart1996implicit,anitescu1997contactLcp, howell2022dojo}. 
As a result, there are many high-performance and mature open-source library implementations that enable reinforcement learning, controller synthesis, and verification in robotics~\cite{todorov2012mujoco,carpentier2019pinocchio,mujocoWarp2026, drake2019}. 

Recently, researchers have leveraged these fast CPU-based simulators to design real-time controllers that achieve impressive results in contact-rich manipulation and locomotion~\cite{zhang2026sumo,zhang2025wholeBodyMpcMujoco,mastalli2020crocoddyl}. Sim-to-real RL has also been accelerated by massively parallel GPU simulation frameworks~\cite{zakka2026mjlab,rudin2022learningWalk}. Similar simulation and control tools are increasingly used for zero-, micro-, and low-gravity robotics~\cite{doerr2023reswarm,spiridonov2024spacehopper}. However, simulators that couple orbit dynamics and external environmental perturbations with existing robotics engines remain underexplored. \mjorbit{} bridges this gap while adding only marginal computational overhead.

\section{Background}
\label{sec:background}

\subsection{Orbital Dynamics and Reference Frames}
\label{sec:bg-orbits}

In this paper, we work with three coordinate systems. The \emph{Earth-centered inertial}
(ECI) frame is non-rotating with its origin at the center of the Earth, and serves as the
primary inertial reference; we use the J2000 realization, with time measured in seconds past
the J2000.0 epoch. Initial states given in other frames are converted to J2000 using the
AstroPy library~\cite{astropyCollaboration2022}. Given a reference orbit with state
$(\bm{r}_c(t),\dot{\bm{r}}_c(t))$ in ECI, we define the \emph{orbit-following} (OF) frame, which
translates with the reference orbit but keeps ECI-parallel axes;
its origin is $\bm{r}_c$ and a body at $\bm{r}_c + \bm{\rho}$ has
OF position $\bm{\rho}$.  The \emph{local-vertical local-horizontal} (LVLH)
frame additionally rotates with the reference orbit: the radial axis points along
$\hat{\bm{r}}_c$, the cross-track axis along $\hat{\bm{r}}_c\!\times\!\dot{\bm{r}}_c$,
and the in-track axis completes the right-handed triad.

A point mass under two-body gravity follows dynamics:
\begin{equation}
  \ddot{\bm{r}} = \bm g_0(\bm{r}),
  \qquad
  \bm g_0(\bm{r}) = -\mu\,\bm{r}/\|\bm{r}\|^3,
  \label{eq:two-body}
\end{equation}
where $\bm g_0$ denotes point-mass gravitational acceleration and $\mu$ is
the central body's gravitational parameter. Practical orbit propagators often
augment $\bm g_0$ with perturbation terms; the leading geopotential correction
for an oblate central body is the standard $J_2$ term, so we write $\bm g$ for
a configured model such as $\bm g=\bm g_0+\bm g_{J_2}$~\cite{vallado2013fundamentals}.
A nearby body at
$\bm{r}_c + \bm{\rho}$ then obeys, in the orbit-following frame,
\begin{equation}
  \ddot{\bm{\rho}}
  = \bm g_0(\bm{r}_c + \bm{\rho}) - \bm g_0(\bm{r}_c)
  + \bm{a}_{\text{ext}},
  \label{eq:relative-motion}
\end{equation}
where $\bm{a}_{\text{ext}}$ collects \emph{specific} (i.e. per-unit-mass) non-gravitational forces.

\subsection{Multibody Dynamics with Contact}
\label{sec:bg-multibody}

An articulated rigid-body system has generalized coordinates
$q\in\mathcal Q$ on a configuration manifold of dimension $n_v$ embedded
in $\R^{n_q}$, where $n_q>n_v$ whenever $q$ carries attitude (typically represented as a unit quaternion).  The generalized velocity
$v\in\R^{n_v}$ is a tangent vector, related to $\dot q$ by a
configuration-dependent map.  The system then satisfies the
manipulator equation, with the applied loads mapped into generalized coordinates by a Jacobian transpose,
\begin{equation}
  M(q)\dot v + \bm{c}(q,v)
  = \bm{\tau}_{\rm act} + J(q)^\top\bm f,
  \label{eq:multibody}
\end{equation}
where $M$ is the mass matrix, $\bm c$ collects Coriolis and centrifugal
terms, and $\bm\tau_{\rm act}$ is direct generalized actuation.  The stacked
wrench $\bm f$ collects every force acting on the bodies---applied body loads,
contact, and joint-constraint forces---and the Jacobian transpose $J(q)^\top$
maps it into a generalized force alongside $\bm\tau_{\rm act}$.
\eqref{eq:multibody} is the form solved by modern robotics engines~\cite{todorov2012mujoco, drake2019,howell2022dojo}.

\section{Methodology}
\label{sec:methods}

\begin{figure}[t]
    \centering
    \input{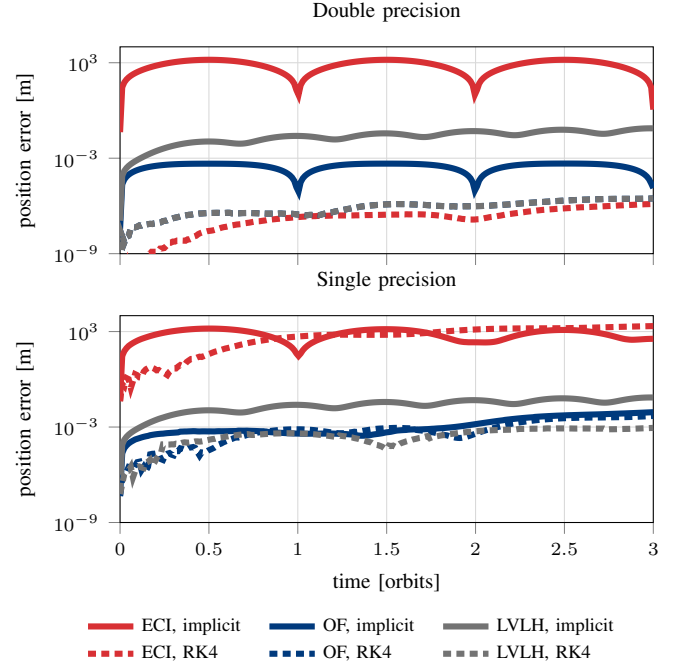}
    \caption{Position errors of an orbit-coupled rigid-body simulation of a single spacecraft under three frame conventions: ECI (red), orbit-following (blue), and LVLH (gray), with implicit (solid) and RK4 (dashed) integrators. RK4 is submillimeter-accurate in all frames in double precision, while the implicit integrator achieves this only in the orbit-following frame. In single precision, only local frames stay accurate over orbital time scales.}
    \label{fig:frame-comparison}
\end{figure}

\subsection{Orbit-Coupled Simulation}
\label{sec:zero-g-comparison}
A common choice when simulating space robots is to simply
disable gravity in an off-the-shelf rigid-body engine and assume the spacecraft lives in a local inertial frame~\cite{schwartz2026smallsatsim}. This drops the relative-gravity term $\bm{g}_0(\bm{r}_c+\bm{\rho}_I)-\bm{g}_0(\bm{r}_c)$ from
\eqref{eq:relative-motion}. This is reasonable over short time scales (minutes) or during Intravehicular Activities (IVA), where the controller counteracts environmental perturbations. Over long external-operation horizons, however, gravitational effects and perturbations accumulate significantly. Figure~\ref{fig:zero-g-comparison}
illustrates this with a head-on cross-track collision between two
free-floating boxes. During and immediately after impact, the trajectories with (solid) and without (dashed) considering gravity are
indistinguishable. However, they noticeably diverge after about $0.2$ hours: in the zero-g baseline, the two boxes drift apart without bound while in the orbit-coupled simulation they oscillate around each other every half orbit.

\subsection{Improving Conditioning with Encke's Method}
\label{sec:encke}
Jointly simulating orbits at thousand-kilometer scales and contact dynamics at meter scales is ill-conditioned and can introduce large integration errors. This is especially important because modern GPU simulators favor single-precision computation. We mitigate these issues in two steps: a cancellation-resistant formulation of differential gravity~\cite{battin1999astrodynamics}, detailed here, and the choice of reference frame for the multibody state, studied in \S\ref{sec:frame-comparison}.

The relative-motion equation~\eqref{eq:relative-motion} keeps the multibody state local to the reference orbit through the differential point-mass gravity term $\bm g_0(\bm{r}_c+\bm{\rho})-\bm g_0(\bm{r}_c)$. Writing the orbit-following position as $\bm{\rho}_I$ to emphasize that it is resolved in ECI-parallel axes, for $\|\bm{\rho}_I\| \ll \|\bm{r}_c\|$ these two gravity vectors nearly cancel. Naively differencing loses ${\sim}\log_{10}(\|\bm{r}_c\|/\|\bm{\rho}_I\|)$ digits of precision --- the regime of meter-scale geometry embedded in a hundred-kilometer-scale orbit.
\rev{In a representative $400$~km LEO scenario ($\|\bm{r}_c\|\approx 6.8\times10^{3}$~km, $\|\bm{\rho}_I\|\approx 1$~m), naive differencing discards about $6.8$ of the roughly $7.2$ significant digits available in single precision, leaving the ${\sim}10^{-6}$~\si{\meter\per\second\squared} differential gravity buried in rounding noise.}
Encke's method~\cite{battin1999astrodynamics, vallado2013fundamentals} rewrites the point-mass difference algebraically as
\begin{equation}
  \bm g_0(\bm{r}_c + \bm{\rho}_I) - \bm g_0(\bm{r}_c)
  = -\frac{\mu}{\|\bm{r}_c\|^3}\bigl(\bm{\rho}_I - f(\sigma)\,(\bm{r}_c + \bm{\rho}_I)\bigr),
  \label{eq:encke}
\end{equation}
where
\begin{align*}
  \sigma
  &= \frac{\bm{\rho}_I\!\cdot\!(2\bm{r}_c+\bm{\rho}_I)}
          {\|\bm{r}_c\|^2},\\
  f(\sigma)
  &= 1 - (1+\sigma)^{-3/2}\\
  &= \frac{\sigma(3 + 3\sigma + \sigma^2)}
          {\left[1+(1+\sigma)^{3/2}\right](1+\sigma)^{3/2}}.
\end{align*}
The final line is the cancellation-resistant form used in our implementation. \eqref{eq:encke} is exact (no Taylor truncation) and never forms the literal subtraction of the two large point-mass gravity vectors. Any formulation whose multibody state is expressed locally about the reference orbit can use this cancellation-resistant gravity. We make this precise in the frame comparison below.

\subsection{Frame Choices for Multibody Simulation}
\label{sec:frame-comparison}
Conditioning the differential gravity removes one source of error; the choice of multibody world frame is the other. We compare three placements: a global ECI frame, a translating OF frame, and a rotating LVLH frame.

An off-the-shelf simulator such as \mj{} treats its world frame as inertial, so any non-inertial-frame acceleration must be supplied externally as a per-body force.
With this correction in place, all three frames from
\S\ref{sec:bg-orbits} are admissible.  ECI carries the absolute position
$\bm{r}_c + \bm{\rho}_I$ and needs no extra terms.  OF carries
$\bm{\rho}_I$ in ECI axes and needs only a translational compensation, the
differential gravity \eqref{eq:relative-motion}.  LVLH carries
$\bm{\rho}_L = C_{LI}\bm{\rho}_I$ in the reference orbit's rotating frame and
additionally requires Coriolis, centrifugal, and Euler-frame terms.  Here
$C_{LI}$ maps inertial components to LVLH components and
$C_{IL}=C_{LI}^\top$.  Concretely, the rigid-body
acceleration integrated in each frame is
\begin{alignat}{2}
  \text{ECI} &:\quad&
  \ddot{\bm{r}}
    &= \bm{g}_0(\bm{r}) + \bm{a}_{\text{ext}},
      \quad \bm{r}=\bm{r}_c+\bm{\rho}_I,
    \label{eq:rhs-eci}\\
  \text{OF} &:\quad&
  \ddot{\bm{\rho}}_I
    &= \bm{g}_0(\bm{r}_c+\bm{\rho}_I)
      - \bm{g}_0(\bm{r}_c) + \bm{a}_{\text{ext}},
    \label{eq:rhs-orbit-following}\\
  \text{LVLH} &:\quad&
  \ddot{\bm{\rho}}_L
    &= C_{LI}\!\left[
        \bm{g}_0(\bm{r}_c+C_{IL}\bm{\rho}_L)
        -\bm{g}_0(\bm{r}_c)
      \right]
      \notag\\
  &&&\quad + C_{LI}\bm{a}_{\text{ext}}
      - 2\bm{\omega}_L\!\times\!\dot{\bm{\rho}}_L
      \notag\\
  &&&\quad
      - \bm{\omega}_L\!\times\!(\bm{\omega}_L\!\times\!\bm{\rho}_L)
      - \dot{\bm{\omega}}_L\!\times\!\bm{\rho}_L,
    \label{eq:rhs-lvlh}
\end{alignat}
where $\bm{\omega}_L$ is the reference orbit's angular velocity expressed in LVLH coordinates. In both the OF and LVLH forms, the differential point-mass gravity is evaluated through the Encke form~\eqref{eq:encke} rather than by literal subtraction, whereas the ECI form integrates the full point-mass acceleration of the absolute position $\bm{r}_c+\bm{\rho}_I$ and cannot use it. While all three are mathematically equivalent, their numerical properties differ in finite precision.

\subsection{Integrator Structure and Symplecticity}
Conditioning alone does not fully determine the best frame. The structure of the equations also matters for the low-order integrators commonly used in real-time and high-throughput rigid-body simulation. In the OF frame, the point-mass acceleration in \eqref{eq:rhs-orbit-following} depends only on position. The translational Hamiltonian remains separable, so the semi-implicit (symplectic) Euler update is symplectic for this component of the dynamics. Its error is therefore typically bounded and oscillatory rather than drifting secularly.

The LVLH equations are local and can use the Encke form for the differential-gravity term, but the rotating-frame terms introduce velocity dependence. In particular, the Coriolis term
$-2\bm{\omega}_L\!\times\!\dot{\bm{\rho}}_L$ breaks the simple separable structure needed by the usual semi-implicit Euler update. ECI has the separable two-body structure, but the multibody engine must carry absolute positions at orbital scale, which is poorly matched to single-precision meter-scale contact simulation. The OF frame is the only one of the three that combines a local coordinate scale, cancellation-resistant differential gravity, and a position-only conservative acceleration.
\begin{figure}[t]
    \centering
    \resizebox{\columnwidth}{!}{%
    \begin{tikzpicture}[
        font=\scriptsize,
        >=Latex,
        box/.style={
            draw,
            line width=1.1pt,
            rounded corners=2pt,
            align=center,
            minimum height=7mm,
            inner sep=3pt,
            text=black
        },
        block/.style={box, draw=plotDarkBlue, fill=plotMedBlue!10},
        sys/.style={box, draw=black!45, fill=black!5},
        state/.style={box, draw=plotGreen!65!black, fill=plotGreen!22},
        phase/.style={box, minimum width=31mm,
            draw=plotOrange!72!black, fill=plotOrange!16},
        flow/.style={->, line width=0.75pt, draw=black!70},
        note/.style={font=\scriptsize, align=center}
    ]
        \node[block, minimum width=24mm] (mjcf) at (-2.7, 4.35)
            {spacecraft\\MJCF};
        \node[block, minimum width=24mm] (cfg) at (0, 4.35)
            {central body\\+ perturbations};
        \node[block, minimum width=24mm] (api) at (2.7, 4.35)
            {actuators +\\sensors};

        \node[sys, minimum width=28mm] (spec) at (0, 3.25)
            {\texttt{MjoSpec}\\MJCF + configs};
        \node[sys, minimum width=28mm] (model) at (0, 2.25)
            {\texttt{MjoModel}\\shared \texttt{mjModel}};
        \node[state, minimum width=38mm] (data) at (0, 1.25)
            {\texttt{MjoData}\(_{1\ldots N}\)\\per-rollout state};
        \node[block, minimum width=22mm] (orbitinit) at (-3.7, 1.25)
            {orbit init};

        \node[phase] (forces) at (0, -0.05)
            {project wrenches\\to generalized loads};
        \node[phase] (mujoco) at (0, -0.95)
            {\mj{} step\\contacts + constraints};
        \node[phase] (advance) at (0, -1.85)
            {advance actuators\\and orbit};
        \node[phase] (cache) at (0, -2.75)
            {refresh environment\\caches};

        \draw[flow] (mjcf) -- (spec);
        \draw[flow] (cfg) -- (spec);
        \draw[flow] (api) -- (spec);
        \draw[flow] (spec) -- node[right, note] {compile} (model);
        \draw[flow] (model) -- node[right, note] {instantiate} (data);
        \draw[flow] (orbitinit) -- (data);
        \draw[flow] (data) -- node[right, note] {\texttt{mjo\_step}} (forces);
        \draw[flow] (forces) -- (mujoco);
        \draw[flow] (mujoco) -- (advance);
        \draw[flow] (advance) -- (cache);
        \draw[flow] (cache.east) -- ++(0.65, 0) |- node[pos=0.26, right, note]
            {updated state} (data.east);

        \node[
            draw=black!55,
            dashed,
            line width=0.7pt,
            rounded corners=2pt,
            fit=(forces) (mujoco) (advance) (cache),
            inner xsep=8pt,
            inner ysep=5pt
        ] {};
    \end{tikzpicture}}
    \caption{\mjorbit{} software pipeline in a familiar \mj{}-inspired API that is shared between CPU and GPU backends.}
    \label{fig:method-overview}
    \vspace{-20pt}
\end{figure}

\subsection{Empirical Frame Comparison}
In Fig.~\ref{fig:frame-comparison}, we simulate a free-floating rigid body in LEO and compare all three frame representations with no additional environmental forces or torques. These placements mirror choices in existing coupled simulators: the ECI frame corresponds to embedding \mj{} as a multibody backend in an inertial world frame~\cite{garciaBonilla2026articulated}, and the OF frame to applying differential gravity at the multibody level, as in MBJEOD~\cite{sullivan2024mbjeod}---though MBJEOD applies this correction only to the articulated bodies and keeps the full system gravity and absolute orbital-scale position on a single root body, rather than placing the entire multibody system in one reference-orbit-following frame. For each frame choice, we integrate rigid-body dynamics with a first-order implicit integrator (implicit) and $4$th-order Runge-Kutta (RK4) integrators with $\Delta t=0.1$s; in frames whose acceleration is position-only (ECI, OF) the implicit update reduces exactly to semi-implicit (symplectic) Euler.
\rev{\mj{} realizes this update as its \texttt{implicit} and \texttt{implicitfast} settings, which differ only in which velocity-derivative terms are treated implicitly and therefore coincide for the position-only dynamics studied here. The full-simulator experiments in \S\ref{sec:mjorbit}--\S\ref{sec:experiments} use \texttt{implicitfast}.} 

We compute the integration error against an independent ground truth obtained by propagating the same body in Basilisk~\cite{kenneally2020basilisk} with its highest-order integrator (Runge-Kutta-Fehlberg 7(8)). In all cases, the orbit dynamics are propagated using RK$4$ with $\Delta t=0.1$s in units of kilometers (km). Rigid-body dynamics simulation uses units of meters (m).
\rev{The step $\Delta t=0.1$~s matches the $10$~Hz control rates of \S\ref{sec:experiments} and the throughput-relevant regime for learning workloads, while still resolving an orbit with ${\sim}5.6\times10^{4}$ steps, keeping RK4 truncation error below the round-off floor. Smaller steps do not help in single precision, because fp32 round-off accumulates with step count and shrinking $\Delta t$ eventually degrades accuracy. Contact-rich examples use the finer \mj{} steps standard for contact (e.g., $10^{-2}$~s).}

In double precision (Fig.~\ref{fig:frame-comparison}, top), RK$4$ maintains accurate simulation through $3$ orbits regardless of frame choice, while the implicit integrator is stable and accurate (sub-millimeter maximum error) only in the OF frame. In single precision (Fig.~\ref{fig:frame-comparison}, bottom), ECI produces large errors even with RK$4$, while the OF frame can produce millimeter-level accurate solutions with just a first-order integrator.
This matches the theory above: ECI suffers from absolute-coordinate scale, LVLH loses the simple symplectic structure through velocity-dependent rotating-frame terms, and OF preserves both local conditioning and symplectic structure. We note that \cite{garciaBonilla2026articulated}, which propagates dynamics with an explicit Runge--Kutta integrator, is closely related to our ECI baseline and exhibits similar absolute-coordinate scaling.

From these experiments, we find the OF frame strictly preferable. It needs no rotational compensation, preserves symplecticity for low-order integrators, and matches the other frames' accuracy under higher-order integration. For the remainder of the paper we default to the implicit integrator and the OF frame. Over explicit semi-implicit Euler, the implicit integrator adds minimal per-step overhead while preserving the rotation-group structure that explicit Euler degrades. In the OF frame its position-only translational update coincides with symplectic Euler, so the bounded error reported here is retained.

\section{\texorpdfstring{\mjorbittitle}{mjorbit}}
\label{sec:mjorbit}
\subsection{Design and Scope}
\mjorbit{} extends \mj{} with orbit state variables, spacecraft actuator and
sensor types, and passive environment forces and torques. This keeps \mj{}'s articulated-body dynamics, constraint solver, MJCF model definition, and control interfaces, adding only the orbit-specific state and coupling needed for space robotics. We build on \mj{} for its extensibility and ecosystem, though the same approach applies to any simulator exposing articulated-body states and generalized force inputs.

\subsection{Reference Orbit and State Representation}
\label{subsec:reference-orbit}
\mjorbit{} couples a central-body orbit propagator to \mj{}'s articulated
rigid-body dynamics. The reference orbit state $(\bm R,\bm V)$ is stored in
a planet-centered inertial frame in multiple standard conventions~\cite{astropyCollaboration2022}; this is the reference (chief) state written $(\bm{r}_c,\dot{\bm{r}}_c)$ in \S\ref{sec:background}--\S\ref{sec:methods}, and each \mj{} body offset $\bm r_i$ below is the corresponding orbit-following (OF)position $\bm{\rho}_I$. \mj{} is in an OF frame described in
\S\ref{sec:frame-comparison}. Translational quantities are converted between SI and km-based orbit units at the \mj{}--orbit boundary; we suppress this conversion below. Thus, for body
$i$ with \mj{} world position and velocity $(\bm r_i,\bm v_i)$,
\begin{equation}
  \bm R_i = \bm R + \bm r_i,\qquad
  \bm V_i = \bm V + \bm v_i,
  \label{eq:absolute-state}
\end{equation}
are the body's absolute inertial position and velocity.

The reference orbit is propagated about a configurable central body under the
gravity model:
\begin{equation}
  \dot{\bm R}=\bm V,\qquad
  \dot{\bm V}=\bm g(\bm R)+\bm a_{\rm fb}.
  \label{eq:reference-dynamics}
\end{equation}
where the feedback acceleration $\bm a_{\rm fb}$ is the mass-weighted average of the non-gravitational accelerations of all the bodies; it is not a physical force, only a choice of moving origin. With total modeled mass $m_{\rm tot}=\sum_i m_i$,
\begin{equation}
  \bm a_{\rm fb} =
  \frac{1}{m_{\rm tot}}\sum_i \bm F_i^{\rm ng,trans},
  \label{eq:feedback}
\end{equation}
where $\bm F_i^{\rm ng,trans}$ contains drag, solar radiation pressure, and
thruster forces.
\rev{Physically, \eqref{eq:feedback} is the non-gravitational acceleration of the system's center of mass, so the origin tracks the mean motion of the modeled system. Without it, sustained thrust, drag, or SRP would drift the multibody state away from the reference orbit and erode the $\|\bm\rho_I\|\ll\|\bm r_c\|$ conditioning of \S\ref{sec:encke}. The matching fictitious force $-m_i\bm a_{\rm fb}$ in \eqref{eq:body-force} cancels it in the absolute state \eqref{eq:absolute-state}. Contact impulses are are excluded as they are internal to the system and cannot accelerate the center of mass.}
\begin{figure}[t]
\vspace{0.05cm}
\centering
\begin{adjustbox}{width=\columnwidth}
\begin{minipage}{0.98\linewidth}
\begin{lstlisting}[language=Python,caption={Minimal \mjorbit{} Python API example.},label={lst:api-example}]
import numpy as np

from mjorbit import MjoModel, OrbitInit, mjo_step
from mjorbit.constants import GM_EARTH, R_EARTH
from mjorbit.testdata import FREE_BODY_XML

# Circular LEO orbit at 400 km altitude.
radius_km = R_EARTH + 400.0
speed_km_s = np.sqrt(GM_EARTH / radius_km)
orbit = OrbitInit(
    R_eci=[radius_km, 0.0, 0.0],
    V_eci=[0.0, speed_km_s, 0.0],
)

# Compile model from MJCF; allocate per-rollout data.
model = MjoModel.from_xml_path(FREE_BODY_XML, mj_timestep=0.01)
data = model.make_data(orbit=orbit)

# Step the coupled orbit and rigid-body dynamics.
mjo_step(model, data)
\end{lstlisting}
\end{minipage}
\end{adjustbox}
\vspace{-0.8cm}
\end{figure}


\subsection{Body Wrenches and Multibody Coupling}
\label{subsec:body-wrenches}

For each \mj{} body, \mjorbit{} assembles an external wrench in the
orbit-following world frame and hands it to \mj{} as a passive load. The translational component is
\begin{equation}
  \bm F_i =
    m_i\!\left[\bm g(\bm R_i)-\bm g(\bm R)\right]
    -m_i\bm a_{\rm fb}
    +\bm F_i^{\rm ng}.
  \label{eq:body-force}
\end{equation}
The first term is differential gravity (point-mass part via the Encke form~\eqref{eq:encke}, $J_2$ difference added when enabled); the second compensates for the origin's non-gravitational acceleration; $\bm F_i^{\rm ng}$ collects body-level environmental and actuator forces. Because the world axes are inertially aligned, no Coriolis,
centrifugal, or Euler-frame terms arise. Drag and solar radiation pressure
(SRP) use standard flat-plate models: per surface, \mjorbit{} rotates the
body-frame normal and center of pressure into the world frame, forms the force
from the projected area against the relative atmospheric velocity and Sun
direction, and adds the resulting center-of-pressure moment to $\bm\tau_i$.
Sun direction and magnetic field are cached at the reference state, whereas
atmospheric density and eclipse are evaluated at each surface's own position,
so bodies offset in altitude or straddling the terminator are handled
correctly.

The rotational wrench $\bm\tau_i$ gathers a gravity-gradient
torque from the body's absolute inertial position and its world-axis inertia
tensor, magnetic torques $\bm\tau=\bm m\times\bm B$ formed in the body frame and
rotated to world coordinates for residual dipoles and magnetorquers, gyroscopic
reaction torques from the stored momentum of reaction wheels and single-gimbal
CMGs, and thruster moments about the body COM. These per-body wrenches $\bm f_i=(\bm F_i,\bm\tau_i)$ stack into the load vector $\bm f$ of~\eqref{eq:multibody} and contribute the passive generalized force
\begin{equation}
  \bm\tau^{\rm passive}
  =J(q)^\top\bm f,
  \label{eq:passive-projection}
\end{equation}
where $J(q)$ stacks the corresponding body Jacobians.
The same wrench formulation therefore applies uniformly to free bodies,
articulated spacecraft, and contact-coupled systems; contact impulses remain
internal to the \mj{} solve and are excluded from the reference feedback
acceleration in~\eqref{eq:feedback}.

\subsection{Simulation Procedure}
\label{subsec:step-procedure}

Both backends carry out the same per-step pipeline shown in
Fig.~\ref{fig:method-overview}: on the CPU it is implemented as a \mj{} engine
plugin attached during model compilation, while the GPU backend realizes the
identical stages as fused MJWarp kernels. A call to \texttt{mjo\_step} executes
four stages:
\begin{enumerate}\setlength{\itemsep}{0pt}
  \item assemble body forces and torques into passive generalized loads via
    \eqref{eq:body-force} and \eqref{eq:passive-projection}, using the
    environment caches for the current reference state;
  \item let \mj{} solve and integrate the constrained multibody dynamics;
  \item advance stored actuator states and propagate the reference state using
    \eqref{eq:reference-dynamics};
  \item refresh the environment caches from the updated reference state for the
    next step.
\end{enumerate}
For comparison with external simulators, \mj{} states convert back to absolute inertial coordinates via~\eqref{eq:absolute-state}.

\begin{figure}[t]
    \centering
    \input{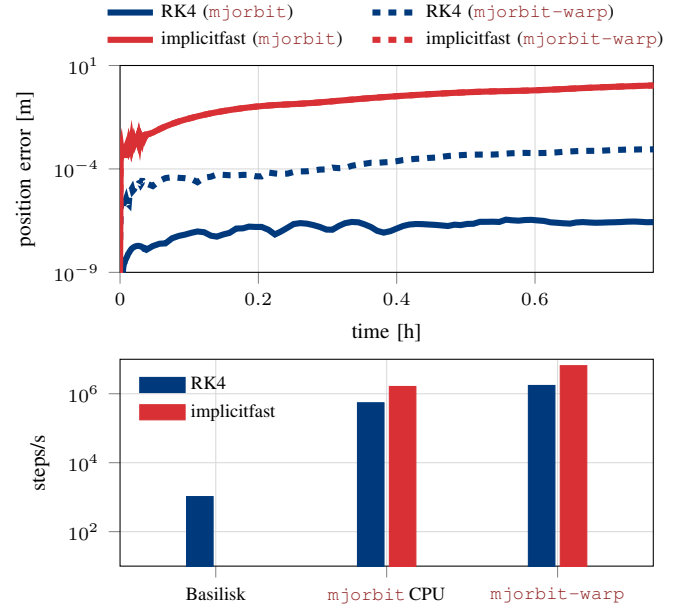}
    \caption{Accuracy and throughput against Basilisk~\cite{garciaBonilla2026articulated} on a bimanual free-flying manipulator in LEO. \textbf{(top)} maximum relative body-position error under RK4 (blue) and implicitfast (red) for \mjorbit{} (fp64, solid) and \mjorbitwarp{} (fp32, dashed). \textbf{(bottom)} both backends deliver orders of magnitude higher throughput than Basilisk.}
    \vspace{-10pt}
    \label{fig:simulation_performance}
\end{figure}

\subsection{API Design and Backends, and Throughput}
The user-facing API follows the same separation as \mj{}: a mutable
specification object is compiled once into an immutable model, while each
rollout owns its own data object. Listing~\ref{lst:api-example} shows the main
objects for an articulated spacecraft with a robotic arm, environment metadata,
and spacecraft actuators. The compiled \texttt{MjoModel} is shared across
threads or batch members, while each \texttt{MjoData} stores the evolving
multibody state, reference orbit, actuator states, sensor buffers, and
per-rollout coupling state.

We provide two backends behind this model/data abstraction. The CPU backend
targets low-latency closed-loop control and runs independent rollouts across
host threads; the GPU backend, \mjorbitwarp{}, builds on MJWarp to execute many
independent worlds in parallel, suiting policy learning, large-scale
evaluation, and dataset generation. Both preserve the same orbit-state convention and per-body coupling, so controllers and experiments move between backends with minimal changes.

\subsection{Trade-offs between Simulation Accuracy and Throughput}

Figure~\ref{fig:simulation_performance} validates both backends against Basilisk
MJScene, an established spacecraft simulator that also embeds \mj{} for
multibody dynamics, on a bimanual free-flying manipulator under matched
smooth-random arm commands, and compares simulation accuracy and throughput. Using the same RK4
integrator, \mjorbit{} on the CPU (double precision) reproduces the Basilisk
reference trajectory to ${\sim}10^{-7}\,\si{\meter}$ relative body-position error
over a half orbit; the single-precision \mjorbitwarp{} path floors near
${\sim}10^{-3}\,\si{\meter}$. The first-order implicitfast integrator instead
drifts to ${\sim}\SI{1}{\meter}$ regardless of precision, so RK4 is required for
high-fidelity attitude over orbital time scales while implicitfast trades
accuracy for speed. On throughput, \mjorbit{}'s native-threaded CPU rollout
sustains ${\sim}0.55$ (RK4) to ${\sim}1.6$ (implicitfast) million steps per
second at eight threads, and \mjorbitwarp{} reaches ${\sim}1.7$ to ${\sim}6.5$
million on an RTX~3080; Basilisk, whose independent MJScene runs are launched
through Python threads, is limited to ${\sim}10^{3}$ steps per second because
its per-scene stepping is serialized by the GIL. \mjorbit{} thus matches Basilisk to numerical precision while running two to three orders of
magnitude faster, with options for accuracy--throughput trade off between RK4 and implicitfast depending on the usecase.

\section{Examples}
\label{sec:experiments}
We demonstrate \mjorbit{} on four on-orbit scenarios (Fig.~\ref{fig:examples}), all in a $\sim$400~km low Earth orbit, exercising both the low-latency CPU backend for real-time control and the GPU backend for large-scale reinforcement learning. The three model-predictive examples share \mjorbit{}'s spline-knot MPPI planner~\cite{howell2022predictiveSampling} in closed loop on the CPU; the learning example trains on the GPU across thousands of parallel worlds.

\subsection{Multibody Attitude Control}
A free-floating bus ($0.8$~m cube, $40$~kg) carries two articulated arms on opposite faces, each a three-degree-of-freedom shoulder plus an elbow ($8$ actuated joints in total); there are \emph{no} bus attitude actuators, so the only way to reorient is to exchange angular momentum with the arms (Fig.~\ref{fig:examples}a). We use an MPPI planner ($5$ spline knots, $128$ rollouts over a $14$~s horizon, replanned at $\sim$3~Hz) to command the arm joint-position setpoints, slewing the bus through a $60^\circ$ rotation about the $(1,1,1)$ axis  and reaching the target in roughly $30$--$40$~s, while a joint-rate penalty keeps the motion gentle and a clearance barrier avoids arm--bus self-collision. The planner then keeps replanning for momentum management, holding attitude against residual J2, drag, and SRP torques.

\subsection{Autonomous Docking}
A $7{,}000$~kg Soyuz-class chaser autonomously docks with a passive $420{,}000$~kg ISS-class target (Fig.~\ref{fig:examples}b). The chaser is fully actuated with three body thrusters ($\pm400$~N) and three reaction wheels ($\pm150$~N$\cdot$m). We use an MPPI planner ($3.5$~s horizon, $128$ rollouts) replanned at $10$~Hz on the CPU to drive the relative position and attitude of the two docking ports to zero, latching once the ports seat (separation $<0.2$~m, attitude error $<5^\circ$) at near-zero relative velocity within a $120$~s episode.
\rev{The capture interface is modeled at the latch level. Hull--hull collision is disabled, and the latch is a conditional weld constraint between the port frames, activated at the thresholds above and enforced as a $0.1$~s critically damped soft constraint. At latch, the model is recompiled with the weld frozen at the current relative pose. Aligned approaches are robust across MPPI settings, while large commanded slews need horizons long enough to brake the maneuver with only ${\sim}0.005$~rad/s$^2$ of wheel authority, and shorter horizons leave the attitude limit-cycling. Frictional multi-point contact is exercised in the Astrobee example below.}

\subsection{Capture and Nadir Pointing}

\begin{figure}[t]
    \centering
    \resizebox{\columnwidth}{!}{\input{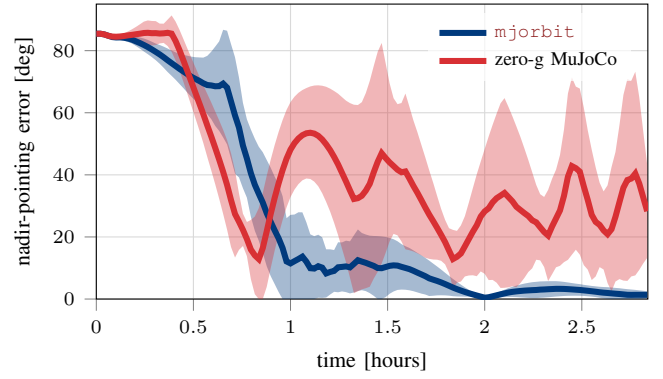}}
    \caption{
    Nadir-pointing error of the captured stack (Fig.~\ref{fig:examples}c) over $5$ trials, with \mjorbit{} (blue) vs.\ zero-g \mj{} (red) as the MPPI rollout dynamics. Orbit-coupled rollouts hold nadir for two hours via the passive gravity-gradient torque, while the zero-g baseline fails.
    }
    \label{fig:mppi-fidelity}
\end{figure}

A free-flying service robot with a slow two-link arm and \emph{no} attitude actuators captures a $400$~kg payload near $90^\circ$ from local vertical (Fig.~\ref{fig:examples}c). It can reorient the captured stack to nadir and hold it only by exploiting the passive gravity-gradient torque through arm motion. We run one MPPI controller whose internal rollouts use either the orbit-coupled \mjorbit{} dynamics or a zero-gravity approximation, and execute \emph{both} on the full \mjorbit{} simulator. Over a two-hour episode and five seeds (Fig.~\ref{fig:mppi-fidelity}), the \mjorbit{}-based planner anticipates the gravity-gradient libration and settles to local vertical on every run with a final-window error $2.2\pm1.2^\circ$, while the planner using zero-g \mj{} cannot and fails on $4$ of $5$ runs ($30.2\pm15.3^\circ$), leaving the stack swinging through nadir. This demonstrates capturing orbital physics that naive simulators ignore directly improves long-horizon task performance.

\subsection{Astrobee Grasping}
An Astrobee robot ($9$~kg) that detumbles from an initial spin, flies to a free-floating cargo module ($5$~kg) drifting $1$--$2$~m ahead, grasps its handle bar with a parallel-jaw gripper, and then station-keeps while holding it (Fig.~\ref{fig:examples}d). We train the policy using the \mjorbitwarp{} on the GPU with $1024$ parallel simulators in $60$~s ($600$-step, $10$~Hz) episodes. The final policy yields an $88\%$ grasp-and-hold success rate across the $1{,}024$ worlds, demonstrating large-scale parallel training for contact-rich on-orbit manipulation.

\section{Conclusions and Limitations}
\label{sec:conclusion}
We have presented a framework for leveraging robotics simulation engines in space robotics settings. Our framework enables accurate long-horizon simulation even under single-precision GPU arithmetic, and we demonstrate it with low-latency CPU and high-throughput GPU backends on a range of space robot control tasks.

\rev{Despite it's promise, several limitations remain in the \mjorbit{} framework. First, within a world, \mjorbit{} inherits \mj{}'s scaling for articulated and contact dynamics, and the orbital layer adds only per-body wrench assembly that is linear in body count. Each world, however, carries a single reference orbit, so the $\|\bm\rho_I\|\ll\|\bm r_c\|$ conditioning of \S\ref{sec:methods} assumes proximity-operations scale. Second, the GPU backend's single-precision clock also loses timing accuracy for time-keyed environment models (sun, eclipse, magnetic field) over multi-hour device-resident rollouts.}

There are several avenues remain for future work. First, beyond the spacecraft sensors it already models (magnetometers, sun sensors), \mjorbit{} still lacks richer perception sensors such as cameras and lidar. \rev{Because the scene remains a standard \mj{} model, \mj{}'s native renderer already produces RGB, depth, and segmentation images of \mjorbit{} states. We plan to expose these as first-class sensors with sun- and eclipse-consistent lighting, along with hooks for external renderers, enabling perception-in-the-loop docking and capture.} Second, this paper only considers first-order implicit and explicit integrators. Future work should investigate variational schemes that better preserve accuracy without the overhead of higher-order integration. Finally, \rev{the validation in this paper is sim-to-sim, and} control policies realized using \mjorbit{} should be validated in hardware tests and on orbit in upcoming missions.
\bibliographystyle{IEEEtran}
\bibliography{references}

\end{document}